\documentclass[acmsmall,screen]{acmart}

\renewcommand\footnotetextcopyrightpermission[1]{}
\AtBeginDocument{%
  }
    
\usepackage{multirow}
\usepackage{bbding}
\usepackage{tabularx}

\usepackage{enumitem}
\usepackage{diagbox}
\usepackage{enumitem}
\usepackage{algorithm}
\usepackage{algorithmic}

\newcommand{\gright}{\textcolor[RGB]{8,156,86}{\Checkmark}}
\newcommand{\rnot}{\textcolor{red}{\XSolidBrush}}

\begin{document}

\title{OPa-Ma: Text Guided Mamba for 360-degree Image Out-painting}

\author{Penglei Gao}
\authornote{Both authors contributed equally to this research.}
\email{gaop@ccf.org}
\affiliation{%
  \institution{Quantitative Health Science Department, Cleveland Clinic}
  \city{Cleveland}
  \state{Ohio}
  \country{USA}
}

\author{Kai Yao}
\authornotemark[1]
\affiliation{%
  \institution{College of Computer Science and Technology, Zhejiang University}
  \city{Zhejiang}
  \country{China}
}

\author{Tiandi Ye}
\affiliation{%
  \institution{East China Normal University}
  \city{Shanghai}
  \country{China}}

\author{Steven Wang}
\affiliation{%
  \institution{Solon High School}
  \country{USA}}

\author{Yuan Yao}
\affiliation{%
  \institution{Tianyuan Biotechnology}
  \country{China}}

\author{Xiaofeng Wang}
\authornote{Corresponding author}
\email{wangx6@ccf.org}
\affiliation{%
  \institution{Quantitative Health Science Department, Cleveland Clinic}
  \city{Cleveland}
  \country{USA}}

\renewcommand{\shortauthors}{Gao et al.}


\begin{abstract}
  In this paper, we tackle the recently popular topic of generating 360-degree images given the conventional narrow field of view (NFoV) images that could be taken from a single camera or cellphone. This task aims to predict the reasonable and consistent surroundings from the NFoV images. Existing methods for feature extraction and fusion, often built with transformer-based architectures, incur substantial memory usage and computational expense. They also have limitations in maintaining visual continuity across the entire 360-degree images, which could cause inconsistent texture and style generation. To solve the aforementioned issues, we propose a novel text-guided out-painting framework equipped with a State-Space Model called Mamba to utilize its long-sequence modelling and spatial continuity. Furthermore, incorporating textual information is an effective strategy for guiding image generation, enriching the process with detailed context and increasing diversity. Efficiently extracting textual features and integrating them with image attributes presents a significant challenge for 360-degree image out-painting. To address this, we develop two modules, Visual-textual Consistency Refiner (VCR) and Global-local Mamba Adapter (GMA). VCR enhances contextual richness by fusing the modified text features with the image features, while GMA provides adaptive state-selective conditions by capturing the information flow from global to local representations. Our proposed method achieves state-of-the-art performance with extensive experiments on two broadly used 360-degree image datasets, including indoor and outdoor settings.
\end{abstract}



\begin{teaserfigure}
\centering
\includegraphics[width=0.95\linewidth]{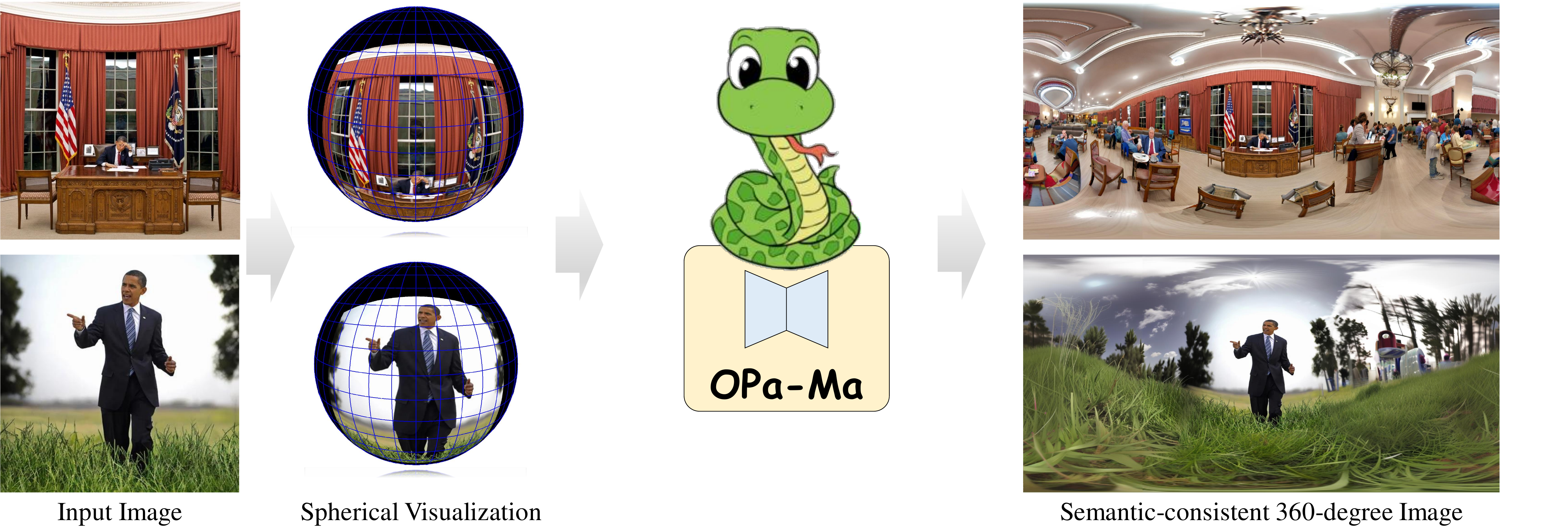}
\caption{We aim to generate smooth and reasonable 360-degree images from NFoV images by utilizing a state space model Mamba with the ability to process long sequences and model spatial continuity.
}
\label{fig:teaser}
\end{teaserfigure}


\maketitle

\section{Introduction}
360-degree image processing has marked a significant evolution in the domain of digital visual content, which could provide a panoramic perspective closely emulating the spherical field of human vision. This advancement facilitates an immersive viewing experience, allowing for a comprehensive exploration of scenes in a manner previously unattainable through conventional imaging techniques. With demanding scenarios, it is an interesting and challenging topic to transform an NFoV image that could be taken from widely available devices into 360-degree visuals, such as a single camera or cellphone. The ability to predict a consistent and plausible extension of these NFoV images into full panoramic views makes the creation of 360-degree content widely accessible, even without the need for specialized panoramic cameras.

\begin{figure*}[ht]
\centering
\includegraphics[width=.95\linewidth]{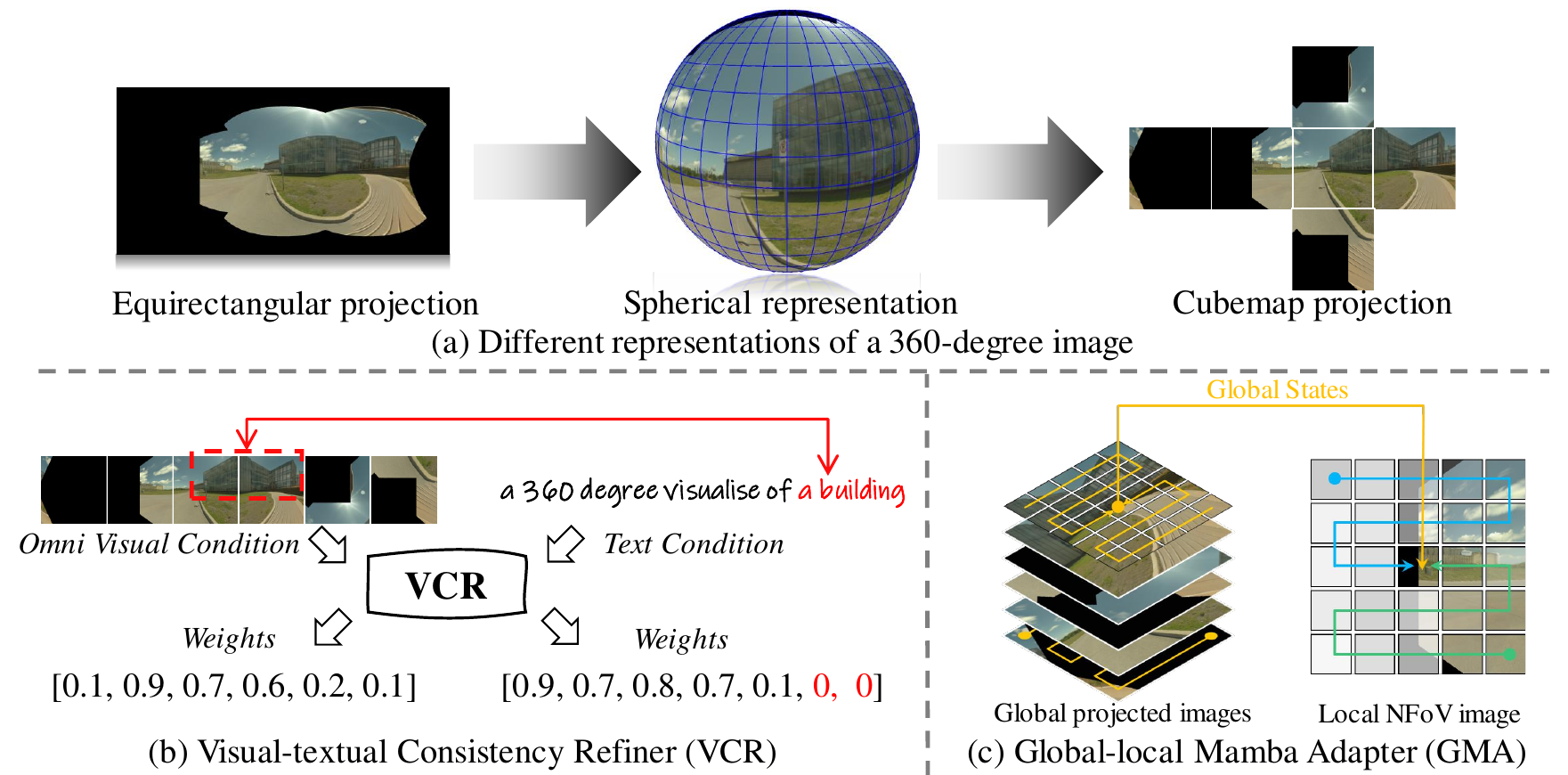}
\caption{Overview of the framework. 
(a) The three different representations of a 360-degree image. (b) The proposed Visual-textual Consistency Refiner (VCR) aims to adjust the image-text condition and provide a better conditional context by comparing the semantic information between the input image and text condition. For example, it will weaken the concept of `a building' in text condition if there already exists information of building in the input image (red dashed box). (c) The proposed Global-local Mamba Adapter (GMA) contributes to connecting the information flow from the global state to local adaptation. The equipped Mamba block demonstrates proficient learning of spatial continuity through bidirectional scanning of a single image.}
\label{fig:banner}
\end{figure*}

With the development of deep learning, many advanced methods have been proposed to solve this interesting and meaningful task, such as convolutional neural networks (CNNs) \cite{sigss} and transformer-based architecture \cite{aognet}. However, the endeavour to accurately generate 360-degree images from NFoV inputs encounters significant challenges. Existing 360-degree out-painting methods often struggle with maintaining visual continuity throughout the panorama, frequently resulting in generated images suffering from inconsistent textures and styles. This discontinuity disrupts the immersive experience, making the expanded views less believable and potentially disorienting. In addition, traditional feature extraction and fusion techniques, particularly those based on transformer architectures, are notably memory-intensive, compounding the challenge of high computational costs and inefficiencies in handling panoramic images' unique spatial characteristics. Furthermore, effective out-painting of 360-degree images requires a deep understanding of the scene context to generate plausible and coherent extensions. Existing methods might lack the sophistication to fully understand complex scenes, leading to generating content that is not contextually appropriate. They are restricted to leveraging sufficient text information to generate more abundant and diverse 360-degree images.

Recently, State Space Models (SSMs) have led to a great application on various machine learning tasks due to their capability of modelling long sequences. 
A sufficient variant of SSM called Mamba~\cite{gu2023mamba} introduces the time-varying updating into the traditional SSM and proposes a work-efficient parallel scanning to enable very efficient training and inference. The success of Mamba demonstrates that the superior scaling ability can be transferred to vision tasks from language modelling. The memory demand for transformer attention mechanisms is quadratic, preventing it from viewing global and local features sequentially within the constraints of available memory. In contrast, Mamba can process the entire image to observe both global and local features with a linear increase in memory consumption. Its unique selection mechanism filters out irrelevant content, retaining only useful information. 
In addition, the capability of modelling spatial continuity \cite{hu2024zigma} could help enhance the extrapolation of unknown regions for 360-degree image out-painting.

In this paper, we propose a novel 360-degree image out-painting method equipped with Mamba (OPa-Ma) to generate abundant and diverse 360-degree panoramic images conditioned on sufficient text guidance. The overall framework is shown in~\autoref{fig:banner}. The capability of processing long sequences and modelling spatial continuity uniquely positions the proposed model to address the intricate task of 360-degree image out-painting. We design two novel modules to process the input images and text conditions. By ingeniously combining image and text features through a novel Visual-textual Consistency Refiner (VCR), our method not only captures but enhances the contextual richness of text guidance. The fusion of modified text features and image features with a weighted sum in the VCR significantly improves the spatial continuity modelling of the entire 360-degree panorama.
Another module called Global-local Mamba Adapter (GMA) is utilized to connect the information flow from the global state to local adaptation. The Mamba adapter processes each NFoV image with the characteristic of selective state space to obtain the global state representation and then captures the local feature of the input NFoV image in a series connection with a global-local Mamba block. The global state representations and local features are obtained to generate more adaptive and efficient conditions for a Diffusion model, facilitating the generation of high-quality panoramic images. Our proposed framework addresses the inherent limitations of existing methods and achieves state-of-the-art performance in the field of 360-degree image out-painting. Extensive experiments conducted on widely recognized 360-degree image datasets, encompassing both indoor and outdoor scenes, underscore the superior capability of our approach to create immersive, high-quality panoramic images from conventional NFoV photographs, heralding a new era in image out-painting technology.

Our contributions can be summarized in the following folders:
\begin{itemize}[leftmargin=*]
    \item We design a Visual-textual Consistency Refiner (VCR) to produce a better conditional context with the input of text guidance and image guidance. The conditional context is obtained with a weighted sum of the modified image feature and text feature with stacked 1D Mamba blocks, providing consistency refining.
    \item We develop a Global-local Mamba Adapter (GMA) to extract the global and local features and connect information flow among the NFoV images with the characteristic of selective state space of Mamba. The local feature is captured of the input image based on the global state representations that are extracted among the multi-direction NFoV images.
    \item Extensive experiments achieve state-of-the-art performance on two widely recognized 360-degree image datasets in indoor and outdoor settings indicating the superiority of our method.
\end{itemize}

\section{Related Work}
\subsection{Image Out-painting}
Image out-painting has emerged as a significant area of interest within the domain of computer vision, aimed at extending visual contexts around a given image with limited known information. Unlike image in-painting, which focuses on filling in missing or damaged parts of images, out-painting seeks to extrapolate reasonable and coherent areas that logically extend the existing scene. With the development of CNNs and Generative Adversarial Networks (GANs) which demonstrate promising ability in generating realistic and contextually appropriate images, many approaches have been proposed based on CNNs to tackle image out-painting \cite{sabini2018painting,van2019image,wang2019wide,yang2019very}.
Based on the work of \cite{yang2019very}, Lu et al.~\cite{lu2021bridging} extrapolated the given image from a new perspective. They aimed to generate vivid content in the intermediate region between two different input images with enhanced contextual information.
Recently, transformers \cite{vaswani2017attention} have been proven to have powerful long-range modelling ability and applied to many vision tasks \cite{LeViT,DeepViT,d2021convit}. Gao et al.~\cite{gao2023generalised} introduced a novel image out-painting method by equipping a transformer-based architecture. 
Yao et al.~\cite{yao2022qotr} developed a patch-wise sequence-to-sequence autoregression framework with a query-based transformer. The input image is divided into several small patches which are further processed with cross-attention considering both the distant and neighbour information of each patch. In the work of~\cite{chang2022maskgit}, the authors built a masked transformer architecture for many image generation tasks including image in-painting, image out-painting, and image manipulation based on the success of auto-encoder-decoder structure of VQGAN~\cite{esser2021taming} model.

\subsection{360-Degree Image Extrapolation}
360-degree image out-painting extends the concept of generalized image out-painting to panoramic images that offer a full spherical view of a scene. This rapidly growing area addresses the unique challenges posed by the spherical format, including the need for seamless integration across the panorama and the handling of spherical distortion. Traditional generalized image out-painting methods often fall short when applied to 360-degree images, given their inability to account for the continuous nature of panoramic scenes and the specific geometric distortions inherent to spherical projection. To tackle this challenging task, many advanced methods have been proposed with different generating behaviours, such as lighting estimation~\cite{gardner2019deep,legendre2019deeplight} and inverse rendering~\cite{sengupta2019neural,wang2021learning}, which indirectly generate the 360-degree images. However, these methods have limitations in generating high-quality images and effectively representing high-frequency textures. In the work of \cite{akimoto2019360}, the authors proposed a GAN-based method for 360-degree image out-painting from pixel completion level and they further developed another diverse transformer-based method \cite{omnidreamer}.
Hara et al. \cite{sigss} proposed a method focusing on scene symmetry to effectively capture the global structure of spherical images from a single NFoV image. Somanath et al. \cite{envmapnet} conducted the 360-degree image out-painting by developing a CNN-based framework, named EnvMapNet. 
Similar to our work, Lu et al.~\cite{aognet} proposed a generative method with text guidance based on stable diffusion~\cite{rombach2022high} for 360-degree image out-painting. They designed two cross-transformer blocks to extract global and local conditioning mechanisms to comprehensively formulate the out-painting guidance with an auto-regressive generation.
However, the above-mentioned methods are restricted to generating smooth 360-degree images with global semantic consistency, which has limitations in modelling spatial continuity.


\subsection{Transformer v.s Mamba}
In the last decades, transformer-based structures have been applied to many fields with diffusion models due to their scalability and effectiveness in kinds of learning tasks \cite{bao2022all,bao2023one,peebles2023scalable,gao2023masked}. The prominent ability to handle long-range dependence and parallel processing makes it more adaptive to many complex data, including images, text and video. 
However, despite the superiority of transformers, there still exist some limitations when applied to diffusion models, such as high memory usage and expensive computational cost. 
Recently, State-Space Models (SSMs) \cite{gedon2021deep,gu2021efficiently,doerr2018probabilistic,fu2022hungry} have been proven to have the promising capability of dealing with long-sequence modelling, and achieve competitive performance to transformer-based methods. Albert Gu and Tri Dao proposed a novel method called Mamba \cite{gu2023mamba} to tackle the issue of transformers’ computational inefficiency on long sequences based on SSM. They improved the SSM by making it a time-variant model to selectively propagate or forget information along the sequence length dimension. Mamba achieves faster inference and linear scaling in sequence length, which enables it to make contributions in various fields. With the emergence of Mamba, variants have been explored in many fields, such as Vision Processing \cite{zhu2024vision,liu2024swin,yang2024plainmamba,hu2024zigma}, Medical Image Analysis \cite{ma2024u,xing2024segmamba}, Graph \cite{wang2024graph,li2024stg} and Time Series \cite{ahamed2024timemachine}.
In this work, we design two modules equipped with both 1D Mamba and 2D Mamba to model the long-range sequence correlation and extract the global to local texture consistency for 360-degree image out-painting.


\section{Preliminary}
\subsection{Mamba} 
State Space Models (SSMs) are designed for continuous systems that map one-dimensional functions or sequences, symbolized as $x(t) \in \mathbb{R}^L \rightarrow y(t) \in \mathbb{R}^L$, by leveraging an intermediate hidden state $h(t) \in \mathbb{R}^N$. In essence, SSMs utilize the subsequent ordinary differential equation (ODE) to represent the input data: 
\begin{align}
h'(t) &= {\mathbf A}h(t) + {\mathbf B}x(t), \\
y(t) &= {\mathbf C}h(t),
\end{align}
where the matrix ${\mathbf A} \in \mathbb{R}^{N\times N}$ is indicative of the system's time evolution, whereas ${\mathbf B} \in \mathbb{R}^{N\times 1}$ and ${\mathbf C} \in \mathbb{R}^{ N\times 1}$ function as projection matrices. The translation of this continuous ODE model into a discrete format is a common practice in contemporary SSMs. A notable discrete variant, Mamba~\cite{gu2023mamba}, introduces a time parameter ${\mathbf \Delta}$ to convert the continuous matrices ${\mathbf A}, {\mathbf B}$ into discrete forms $\overline{{\mathbf A}}, \overline{{\mathbf B}}$. This conversion commonly utilizes the zero-order hold (ZOH) method, described by: 

\begin{align}
\overline{{\mathbf A}} &= \exp({\mathbf \Delta \mathbf A}),\quad
\overline{{\mathbf B}} = ({\mathbf \Delta \mathbf A})^{-1} (\exp({\mathbf \Delta \mathbf A}) - {\mathbf I}) \cdot {\mathbf \Delta \mathbf B}, \\
h_t &= \overline{{\mathbf A}} h_{t-1} + \overline{{\mathbf B}} x_t,\quad
y_t = {\mathbf C}h_t.
\end{align}

\begin{figure}[t]
\centering
\includegraphics[width=0.7\linewidth]{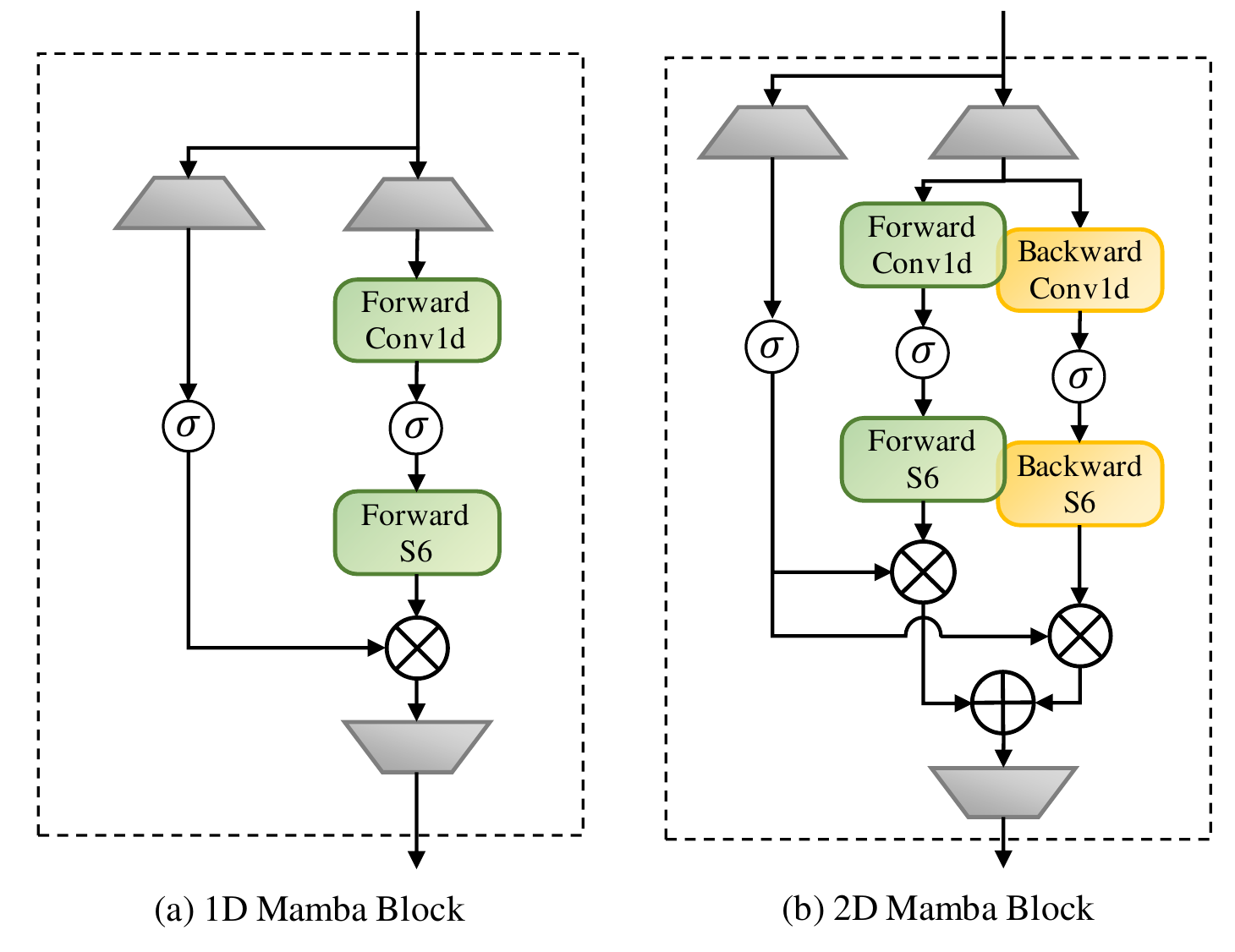}
\caption{Mamba block for 1D sequential tasks and 2D visual tasks.}
\label{fig:mamba}
\end{figure}

Unlike typical models that depend on linear and time-invariant SSMs, Mamba introduces a selective scan mechanism (S6)  for long-sequence modelling. The S6 method determines the parameters ${\mathbf B} \in \mathbb{R}^{B\times L \times N}$, ${\mathbf C} \in \mathbb{R}^{B\times L \times N}$, and ${\mathbf \Delta} \in \mathbb{R}^{B \times L \times D}$ directly from the data input $x \in \mathbb{R}^{B \times L \times D}$, showcasing a profound ability for contextual awareness and dynamic modulation of weights. \autoref{fig:mamba}(a) details the Mamba block for 1D sequence modeling. However, the original Mamba, although adept at handling 1D sequences, falls short in addressing the demands of visual tasks that require spatial awareness. In response to this limitation, Vision Mamba~\cite{zhu2024vision} introduces a bidirectional 2D Mamba block designed specifically for bidirectional sequence processing within visual contexts, as shown in~\autoref{fig:mamba}(b). This innovative block processes flattened visual sequences using simultaneous forward and backward S6 blocks, considerably improving its capability for spatial understanding.

\begin{figure*}[t]
\centering
\includegraphics[width=0.9\linewidth]{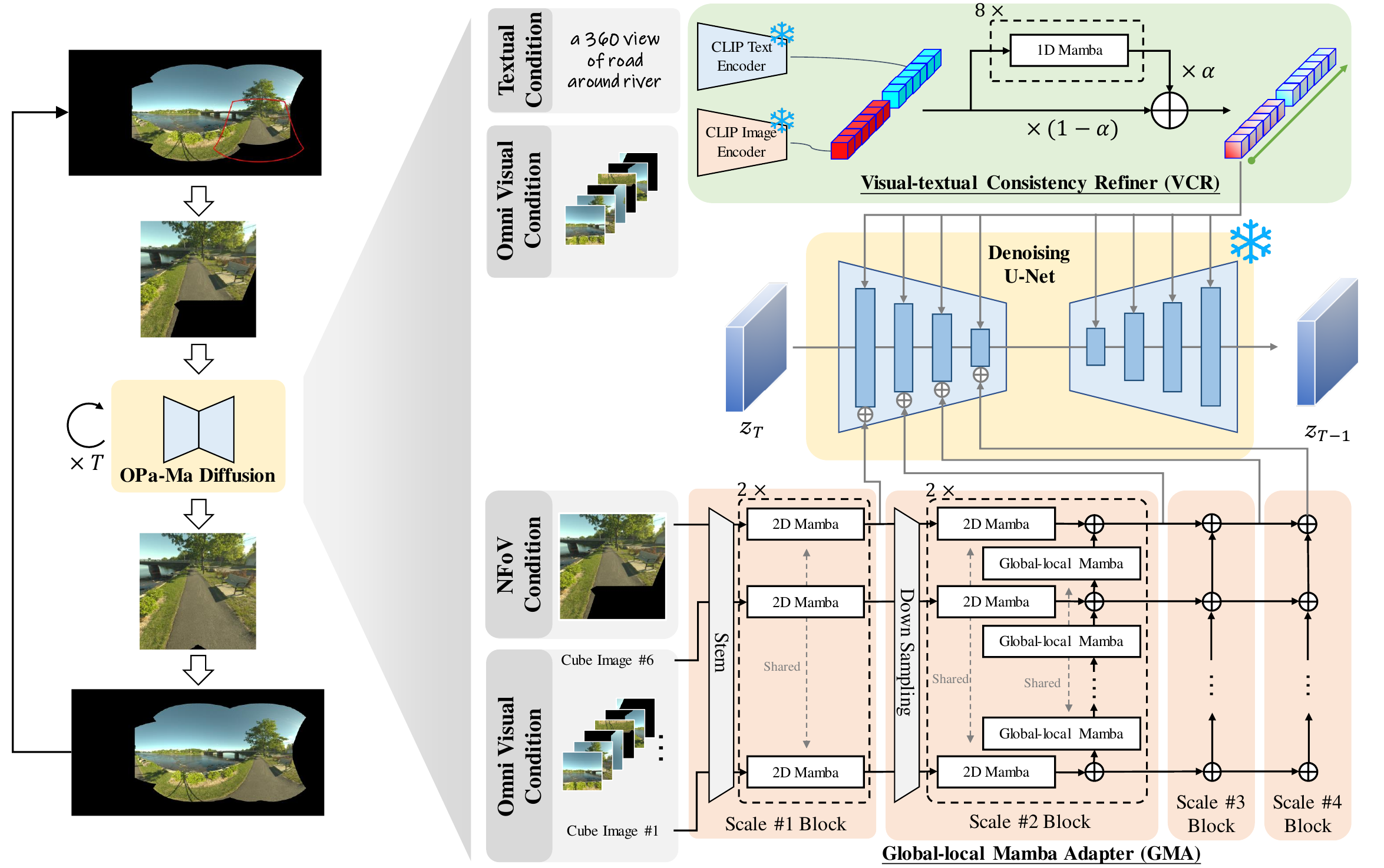}
\caption{Architecture of the proposed OPa-Ma. The left part is the overall generation scheme which will generate panorama images iteratively with OPa-Ma Diffusion. The right part is the detail of OPa-Ma Diffusion.
VCR and GMA could provide a better condition for Denoising U-Net from two perspectives. 
Stacked 1D Mamba blocks are utilized to obtain the modified image-text features extracted by the pre-trained CLIP model. We use a re-weight mechanism to achieve consistency refining. In the GMA module, the NFov input image and the omni visual condition will be processed by shared 2D Mamba individually. The global-local Mamba controls the information flow and extracts the local features of the cube images one by one with the state-selective characteristics of Mamba. There are four scale blocks in GMA and each one outputs a targeted condition for the encoder of Denoising U-Net.}
\label{fig:architecture}
\end{figure*}

\subsection{Latent Diffusion Model} 
In this paper, we consider latent diffusion models~\cite{rombach2022high} as a generation method, consisting of an autoencoder and diffusion model. The encoder $\mathcal{E}$ encodes image $x$ into a latent representation $z=\mathcal{E}(x)$, and the decoder $\mathcal{D}$ reconstructs the image from latent, giving $\widetilde{x}=\mathcal{D}(\mathcal{E}(x))$. Given an image $x_0$, we project it into latent $z_0=\mathcal{E}(x_0)$, and apply a forward diffusion Markov process to add noise across a series of time steps $t$ with scheduled variance $\beta_t$:
\begin{align}
    q(z_t|z_{t-1}) &= \mathcal{N}\left(\sqrt{1-\beta_t} x_t, \beta_t \mathbf{I}\right)\\
    q(z_{1:T}|z_0) &= \prod q(z_t|z_{t-1}). \nonumber
    \label{eq:forward}
\end{align}
$T$ is the total number of steps, $\mathbf{I}$ is the identity matrix. If $T\rightarrow\infty$, the output $z_T$ will be isotropic Gaussian. The specific definition of Markov chain facilitates the derivation of $z_t$ in a closed form:

\begin{align}
    z_t &= \sqrt{{\alpha}_t}z_{t-1} + \sqrt{1-{\alpha}_t}\epsilon_{t-1}= \sqrt{\bar{\alpha}_t}z_0 + \sqrt{1-\bar{\alpha}_t}\epsilon,
\end{align}
where $\alpha_t=1-\beta_t$, $\bar{\alpha}_t=\prod_{i=1}^t \alpha_i$, $\epsilon_t \sim \mathcal{N}(0, \mathbf{I})$.

To synthesize images from random noise, we trained a neural network $p_{\theta}$ to approximate the conditional distribution $q(z_{t-1}|z_t)$, which also behaves as a Gaussian when $\beta_t$ is sufficiently small. 
\begin{align}
    p_{\theta}(z_{t-1}|z_t) &= \mathcal{N}\left(\mu_{\theta}(z_t, t), \Sigma_{\theta}(z_t, t)\right),
\end{align}
where $\mu_{\theta}$ is trained to predict $z_{t-1}=\frac{1}{\sqrt{\alpha_t}}\left(z_t - \frac{1-\alpha_t}{\sqrt{1-\bar{\alpha}_t}\epsilon_t}\right)$, which is derived from \autoref{eq:forward}.
Since we already have $z_t$ during training, a network $\epsilon_{\theta}$ is trained to predict $\epsilon_t$ instead of $\mu_{\theta}$ \cite{ho2020denoising}, facilitating the attainment of the training objective for the latent diffusion model: 
\begin{align}
    \mathcal{L} &= \mathbb{E}_{t\sim [1, T], z_0, \epsilon_t}\left\|\epsilon_t - \epsilon_{\theta}(z_t, t) \right\|_2^2
\end{align}

At test time, we start from a random noise $z_T\sim \mathcal{N}(0, \mathbf{I})$ and then iteratively apply the model $\epsilon_{\theta}$ to obtain $z_{t-1}$ from $z_t$ until $t=0$.
Then, we can reconstruct images from latent representation using decoder, i.e., $x_0=\mathcal{D}(z_0)$.

In conditional latent diffusion models, conditional information can be seamlessly integrated into the $\epsilon_{\theta}$ network without changing the loss function. This enables the model to effectively leverage the provided conditions to generate high-quality conditional images.

\section{Method}
Given a 360-degree image $X$, there are three types of representations which can be transformed into others, including equirectangular projection, spherical representation, and cubemap projection, as illustrated in \autoref{fig:banner}(a). Equirectangular projection is adopted as the standard form for a 360-degree image, where $X \in \mathbb{R}^{H\times W \times 3}$. From the Spherical representation, one can extract any desired local Normal Field of View (NFoV) image using specific longitude $\omega \in [-90^\circ,90^\circ]$ and latitude $\phi \in [0^\circ, 180^\circ]$, denoted as  $x = X(\omega, \phi) \in \mathbb{R}^{H_c\times W_c \times 3}$. Cubemap projection maps $X$ into the faces of a cubic, where $X=\{x_F,x_L,x_B,x_R,x_U,x_D\} \in \mathbb{R}^{H_c\times W_c \times 3}$ can be viewed as a general NFoV image. In this work, we aim to outpaint a local NFoV image $x$ with a text prompt $\mathcal{T}$ to get a 360-degree image $X$. 

As depicted in \autoref{fig:architecture}, our proposed OPa-Ma framework generates 360-degree images by progressively out-painting local regions based on latent diffusion methods. In each step, a local NFoV image is extracted from an incomplete 360-degree image, encompassing both adjacent unknown and known areas. Meanwhile, various conditions are introduced to guide the diffusion process, including omni visual condition, textual condition, and NFoV condition. Two modules are developed to better leverage various conditions for more accurate image generation, including the Visual-textual Consistency Refiner (VCR) and the Global-local Mamba Adapter (GMA). We will detail these two modules in the following sections.



\subsection{Visual-textual Consistency Refiner}
To facilitate text-guided omni-image out-painting, OPa-Ma considers both global omni-visual conditions and textual conditions of a single local NFoV image. Typically, the input text guidance is intended for the entire image, making it suboptimal as a guidance for a local image. Additionally, global visual features can sometimes conflict with textual conditions, resulting in unreasonable outputs. This motivates us to generate a better conditional context for the extrapolation of local NFoV images based on global conditions, including both global visual conditions and textual conditions.

We leverage a pre-trained CLIP model to extract the global context. Specifically, we use a text encoder $\mathcal{E}_{text}$ for retrieving textual information, formulated as $c_{text}=\mathcal{E}_{text}(\mathcal{T})$, and an image encoder $\mathcal{E}_{image}$ to derive global visual features from cubemap projection of NFoV images, represented as $c_{omni}=\{\mathcal{E}_{image}(x_F),\dots,\mathcal{E}_{image}(x_D)\}$ $\in \mathbb{R}^{6\times d}$, with $d$ indicating the hidden dimension of CLIP. It is noteworthy that texts are padded to the length of 77, ensuring a fixed dimension for $c_{text} \in \mathbb{R}^{77\times d}$. These components are then combined to formulate the raw clip condition $c_{clip}=\{c_{omni},c_{text}\} \in \mathbb{R}^{83\times d}$. 
To achieve better semantic consistency, we use mamba blocks for sequence-to-sequence processing. In particular, global semantic features are processed by Mamba blocks, i.e., $z_{clip} = Mamba(c_{clip}+ E )$, where $ E \in \mathbb{R}^{83\times d}$ are learnable positional embeddings. Then, the reweighted semantic condition is formatted as follows:
 \begin{align}
    c_{vcr} &= \alpha \times c' + (1-\alpha) \times c_{clip} ,\\
    c'  =  &h_1(z_{clip}), \alpha  = \texttt{Sigmoid}(h_2(z_{clip})),
\end{align}
where $h_1$ and $h_2$ are two linear heads.  In the training process, we randomly replace 50\% text prompts $\mathcal{T}$ with empty strings, enabling us to out-paint without text guidance.
With the proposed VCR, we can adjust the text information based on global visual information to achieve better overall semantic consistency.


\subsection{Global-local Mamba Adapter}
For omni image out-painting, we propose the GMA module to simultaneously process and merge visual information from global to local conditions. Specifically, a local NFoV image $x$ is used alongside Cubemap-projected general NFoV images $\{x_F,x_L,x_B,x_R,x_U,x_D\}$ as input. These inputs combine the geometry's coordinates with the image channels before being fed into the adapter.  Image features are then extracted across four resolution scales, $\{ \frac{1}{8},\frac{1}{16},\frac{1}{32},\frac{1}{64}\}$, to align with the latent diffusion model. At each scale, 2D mamba blocks are utilized to extract information within an individual image. Meanwhile, for scale 2$\sim$4, global-local mamba blocks are introduced to facilitate the transfer of information among images through state passage, achieving enhanced information fusion of global and local context. 
Finally, multi-scale fused local texture condition features $c_{gma}=\{c_{gma}^1,c_{gma}^2,c_{gma}^3,c_{gma}^4 \}$ are formed.

\begin{table*}[ht]
\small
\centering
\caption{Comprehensive comparison with the state-of-the-art 360-degree image out-painting methods.}
\label{quan_overall}
\setlength\tabcolsep{4 pt}
\begin{tabular}{cc|cccccc}
\toprule[1pt]
NFoV Image Guidance & Text Guidance & Method & Venue & \multicolumn{2}{c}{Indoor Dataset} & \multicolumn{2}{c}{Outdoor Dataset} \\\midrule
\multirow{6}{*}{\gright} & \multirow{6}{*}{\rnot} &  &  & FID~$\downarrow$ & LPIPS~$\downarrow$ & FID~$\downarrow$ & LPIPS~$\downarrow$ \\
 &  & SIG-SS~\cite{sigss} & AAAI'21 & 197.4 & - & - & - \\
 &  & EnvMapNet~\cite{envmapnet} & CVPR'21 & 52.70 & - & - & - \\
 &  & OmniDreamer~\cite{omnidreamer} & CVPR'22 & 46.15 & 0.45 & 24.50 & 0.41 \\
 &  & AOG-Net~\cite{aognet} & AAAI'24 & 38.60 & 0.37 & 18.40 & \textbf{0.36} \\
 &  & OPa-Ma & - & \textbf{9.58} & \textbf{0.36} & \textbf{17.50} & \textbf{0.36} \\ \midrule
\multirow{4}{*}{\rnot} & \multirow{4}{*}{\gright} &  &  & SC~$\uparrow$ & IS~$\uparrow$ & SC~$\uparrow$ & IS~$\uparrow$ \\
 &  & Text2Light~\cite{textlight} & TOG'22 & 0.33 & 4.5 & 0.45 & 3.9 \\
 &  & AOG-Net~\cite{aognet} & AAAI'24 & 0.36 & 5.1 & 0.53 & 4.2 \\
 &  & OPa-Ma & - & \textbf{0.68} & \textbf{6.4} & \textbf{0.69 }& \textbf{4.3} \\ \midrule
\multirow{4}{*}{\gright} & \multirow{4}{*}{\gright} &  &  & FID~$\downarrow$ & LPIPS~$\downarrow$ & FID~$\downarrow$ & LPIPS~$\downarrow$ \\
 &  & ImmenseGAN~\cite{immersegan} & 3DV'22 & 42.78 & - & - & - \\
 &  & AOG-Net~\cite{aognet} & AAAI'24 & 9.76 & 0.39 & 18.80 & 0.41 \\
 &  & OPa-Ma & - & \textbf{7.60} & \textbf{0.35} & \textbf{13.28 } & \textbf{0.34} \\ \bottomrule[1pt]
\end{tabular}
\end{table*}

\begin{figure*}[t]
\centering
\includegraphics[width=.95\linewidth]{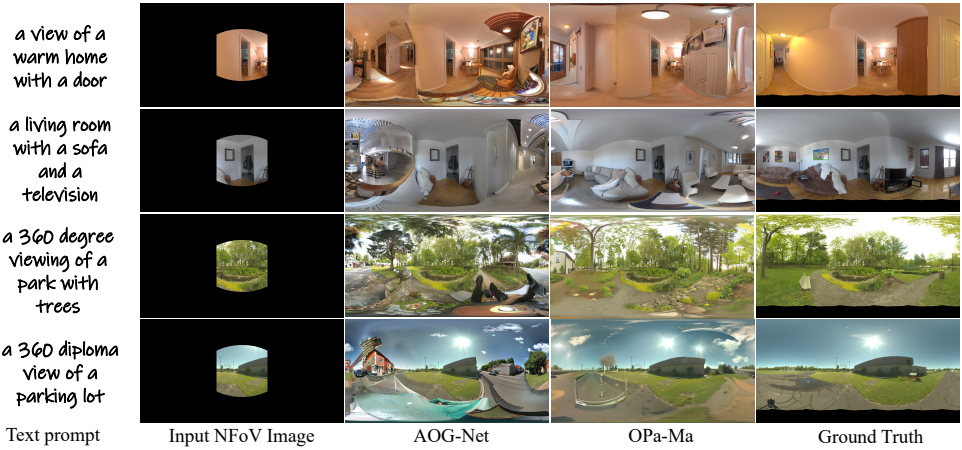}
\caption{Visual results on indoor and outdoor settings with both NFoV image and text guidance. 
}
\label{fig:both}
\end{figure*}


\subsection{Model Optimization}
During optimization, we fix the parameters in the latent diffusion model and CLIP, while optimising the proposed VCR and GMA. Each training sample is a triplet, containing an incompleted 360-degree image $X$, local view coordinates $(\omega,\phi)$, and text prompt $\mathcal{T}$. First, the local view NFoV image $x_0=X(\omega,\phi)$ is embedded to latent representation $z_0$ via the image encoder. Then, we randomly sample a time step $t$ from $[0, T]$ and add corresponding noise to $z_0$, producing $z_t$. Last, the network is optimized via:
\begin{align}
    \mathcal{L} &= \mathbb{E}_{t\sim [1, T], z_0, \epsilon_t}\left\|\epsilon_t - \epsilon_{\theta}(z_t, t, c_{vcr}, c_{gma}) \right\|_2^2
\end{align}

\section{Experiments}
\subsection{Experimental Setup}
\noindent \textbf{Datasets}
In alignment with previous research~\cite{envmapnet,omnidreamer,aognet}, the evaluation is performed using the LAVAL indoor HDR dataset~\cite{DBLP:journals/tog/GardnerSYSGGL17} and the LAVAL outdoor HDR dataset~\cite{DBLP:conf/iccv/ZhangL17}.
For indoor settings, the dataset contains 2,233 spherical indoor images, offering comprehensive coverage of varied indoor environments, each with a resolution of 7,768 $\times$ 3,884. There are 1,921 training images and 312 testing images according to the official split.
As for the outdoor dataset, it contains 205 high-resolution images. 
A training subset of 165 images is selected, with the remaining 40 serving as the test set based on \cite{aognet}.
In the processing stage for both indoor and outdoor datasets, image resolution is reduced to 4,096 by 2,048 pixels to optimize computational efficiency during training.

Due to the lack of paired text caption and image in both datasets, we utilized the advanced captioning model BLIP2~\cite{DBLP:conf/icml/0008LSH23} to automatically produce textual descriptions for the 360-degree image following \cite{aognet}.
An initial caption is generated for each image in its equirectangular projection, providing concise overviews. Subsequently, we create additional captions for the individual horizontal faces of the cubemap, further enriching the textual context.


\begin{figure*}[t]
\centering
\includegraphics[width=.95\linewidth]{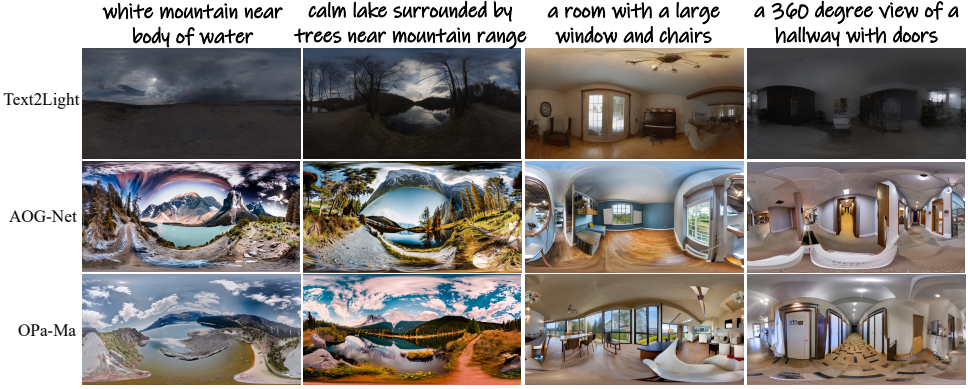}
\caption{Visual results with only text guidance on indoor and outdoor settings.
}
\label{fig:text}
\end{figure*}

\begin{figure*}[t]
\centering
\includegraphics[width=.95\linewidth]{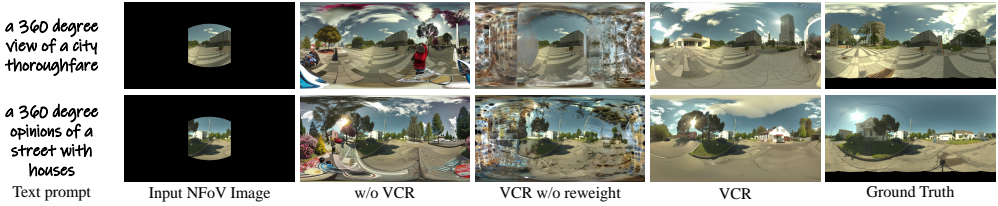}
\caption{Effect of Visual-textual Consistency Refiner (VCR) in OPa-Ma. }
\label{fig:vcr}
\end{figure*}

\noindent \textbf{Implementation Details}
In our experiments, we utilize the pre-trained Stable Diffusion \cite{rombach2022high} as the latent diffusion model, and pre-trained OpenCLIP \cite{cherti2023reproducible} for text encoder $\mathcal{E}_{text}$ and image encoder $\mathcal{E}_{image}$. Eight stacked 1D Mamba blocks are used in the VCR.
OPa-Ma is trained using AdamW optimizer with learning rate 1e-4 and batch size 1. For inference, DPM-Solver++~\cite{lu2022dpm} is employed as a sampler with step size 25 and a classifier-free-guidance scale set to 2.5. All experiments are conducted on an NVIDIA RTX 4090.

\noindent \textbf{Evaluation Metrics}
To evaluate the performance of our method, we utilize LPIPS~\cite{DBLP:conf/cvpr/ZhangIESW18} and Fréchet Inception Distance (FID)~\cite{DBLP:conf/nips/HeuselRUNH17} as the quantitative metrics to measure the generative quality. 
We adopt the advanced captioning model BLIP2~\cite{DBLP:conf/icml/0008LSH23} to assess Semantic Consistency (SC), where the estimated caption is derived from the generated image. SC is calculated using sentence embeddings~\cite{DBLP:conf/emnlp/ReimersG19} between the estimated caption and the input text.
Furthermore, the Inception Score (IS)~\cite{DBLP:conf/nips/SalimansGZCRCC16} is used as an additional indicator to measure the generated image's quality.



\subsection{Comparison with State-of-the-Art}
\noindent \textbf{Baselines}
We evaluate the performance of our method from three experimental settings according to the types of input conditions with several state-of-the-art Omni out-painting methods. (1) \textit{Without text guidance}: The text condition is set as a blank prompt only with NFoV image input. We compare the performance of FID and LPIPS with SIG-SS~\cite{sigss}, EnvMapNet~\cite{envmapnet}, OmniDreamer~\cite{omnidreamer}, and AOG-Net~\cite{aognet} for this setting. (2) \textit{Without NFoV image guidance}: For the text-only generation setting, the initial NFoV image is generated with a pre-trained SD model. AOG-Net~\cite{aognet} and Text2Light~\cite{textlight} are compared for evaluation on SC and IS metrics. (3) \textit{With both NFoV image and text guidance}: We compare the evaluation metrics of FID and LPIPS with two methods, ImmenseGAN~\cite{immersegan} and AOG-Net~\cite{aognet}. 

\noindent \textbf{Quantitative Results}
\autoref{quan_overall} shows the overall performance on the three experimental settings. Note that SIG-SS, EnvMapNet, and ImmenseGAN didn't evaluate outdoor settings in the literature and we only report the results of indoor settings. 
With only NFoV image guidance, our method achieves a great improvement on the Indoor dataset compared with the most powerful method AOG-Net with an FID score 9.58 and an LPIPS value 0.36. This result shows that our method is generalized and robust with input image and text conditions. Although training with both NFoV image guidance and text guidance, our method could generate high-quality 360-degree images during inference when the text condition is set as a blank prompt. Regarding the comparison of only involving text guidance in the second part of \autoref{quan_overall}, our method outperforms AOG-Net and Text2Light with promising quantitative results with an SC score 0.68 and an IS value 6.4 under the indoor setting and an SC score 0.69 and an IS value 4.3 under the outdoor setting. The proposed OPa-Ma achieves a high semantic consistency even on the more complex indoor dataset which lacks in-depth text description. In the third part of \autoref{quan_overall}, we show the model performance compared with ImmenseGAN and AOG-Net, which conduct the 360-degree image out-painting with both NFoV image guidance and text guidance. The results demonstrate the superiority of our method with a great decrease in FID score and LPIPS value for the outdoor dataset.

\begin{table}[t]
\small
\centering
\caption{Ablation study on VCR and GMA.}
\label{tab:module}
\begin{tabular}{cc|cc}\toprule[1pt]
\multicolumn{2}{c|}{Method} & \multicolumn{2}{c}{Outdoor Dataset} \\
VCR & GMA & FID$\downarrow$ & LPIPS$\downarrow$ \\ \midrule
- & - &32.22  &  0.42 \\
- & \gright & 24.99 & 0.42 \\
\gright & - & 14.63 & 0.35 \\
\gright & \gright & \textbf{13.28} & \textbf{0.34} \\\bottomrule[1pt]
\end{tabular}
\end{table}

\begin{figure}[t]
\centering
\includegraphics[width=.9\linewidth]{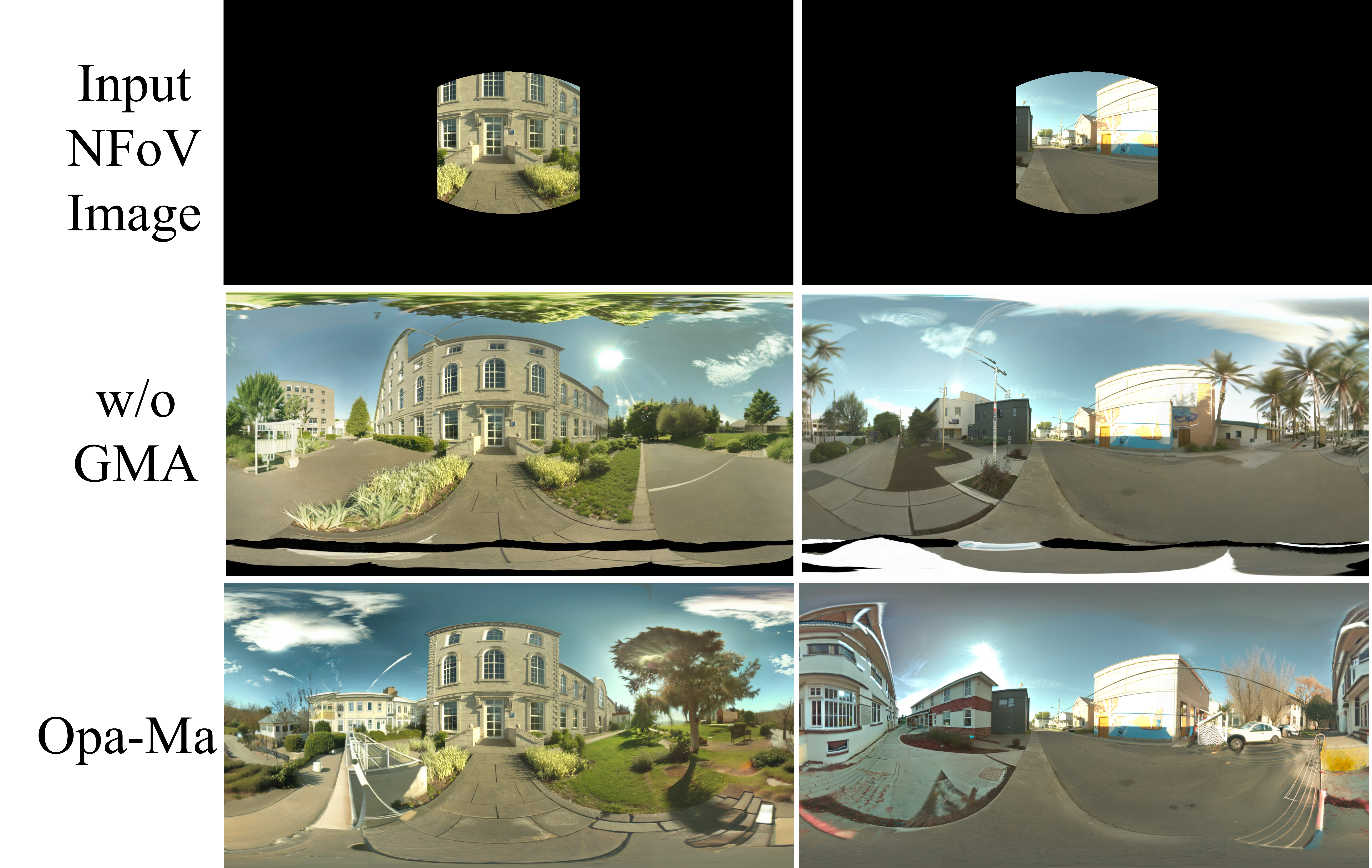}
\caption{Visual results on the effect of GMA.}
\label{fig:gma}
\vspace{-0.5em}
\end{figure}

\noindent \textbf{Qualitative Results}
In \autoref{fig:both}, we present the visual results with both NFoV image and text guidance compared with AOG-Net. In the 1st row of \autoref{fig:both}, our method could generate complete and consistent walls and doors in the house, while AOG-Net suffers from obtrusive content on the right side and bottom of the generated image. In the 2nd row, our method is able to produce the required sofas and generate a reasonable panorama of the living room. In addition, the view of a park is more smooth and realistic generated by our method OPa-Ma in the 3rd row, with detailed trees and coherent ground. The generated image contains several bad pixels and a person of AOG-Net, which will affect the overall coherence and smoothness of the image. To evaluate the generative ability with text guidance, we show several samples only with text guidance in \autoref{fig:text}. Compared with Text2Light, our method could generate visually appealing 360-degree images, while the generated images are more monotonous and lack of details, leading to degenerated visual quality. Furthermore, the generated 360-degree images are more smooth and realistic with vivid details of OPa-Ma. On the contrary, the generated contents lack spatial continuity in the edge of the images of AOG-Net. For example, the ground and sky near the lake are inconsistent and there are several unexpected white blocks on the floor respectively in the 2nd and 4th columns in~\autoref{fig:text}.

\subsection{Analytic Study}

\noindent \textbf{Effect of Visual-textual Consistency Refiner}
We show the quantitative results of the effect of module VCR in \autoref{tab:module}. Without VCR, the FID score decreases much and the LPIPS value gets worse. In~\autoref{fig:vcr}, we can see that the generated images contain many redundant contents with inconsistent semantic information. These results indicate that the text prompt is difficult to take effect and the model cannot maintain global semantic consistency without VCR. When there is no re-weight mechanism in VCR, the model is restricted for convergence, which could be visually presented in \autoref{fig:vcr}. The re-weight mechanism helps to keep the global textual semantic consistency and generate visually appealing images.


\noindent \textbf{Design of Global-local Mamba Adapter}
Although the FID score and LPIPS value are similar to the full model in~\autoref{tab:module}, our method could generate complete and detailed 360-degree images with module GMA shown in~\autoref{fig:gma}. The generated images have corrupt regions with black and white pixels without GMA, which could be caused by lacking the ability to model spatial continuity. In~\autoref{tab:magic}, we present the effectiveness of global-local fusion in GMA. The results show that our method achieves the best performance when adding global-local Mamba in 2,3 and 4-th layers of Denoising U-Net to obtain the best balance of efficiency and performance. The performance gets better when gradually increasing the influence of global-local Mamba from deep to shallow layer. However, the fine-grained fusion of global-local information could cause a decline in model performance for all layers. 

\begin{table}[t]
\small
\centering
\caption{Effect of GMA, where * indicate the default setting.}
\label{tab:magic}
\begin{tabular}{ccc}\toprule[1pt]
\begin{tabular}[c]{@{}c@{}}Scales with \\ Global-local Mamba\end{tabular} & FID$\downarrow$ & LPIPS$\downarrow$ \\\midrule
$[4]$ & 17.77 & 0.35 \\
$[3, 4]$ & 14.59 & 0.35 \\
$[2, 3, 4]*$ & \textbf{13.28} & \textbf{0.34} \\
$[1,2,3,4]$ & 15.36 & 0.36\\\bottomrule[1pt]
\end{tabular}
\end{table}

\begin{figure}[t]
\centering
\includegraphics[width=.8\linewidth]{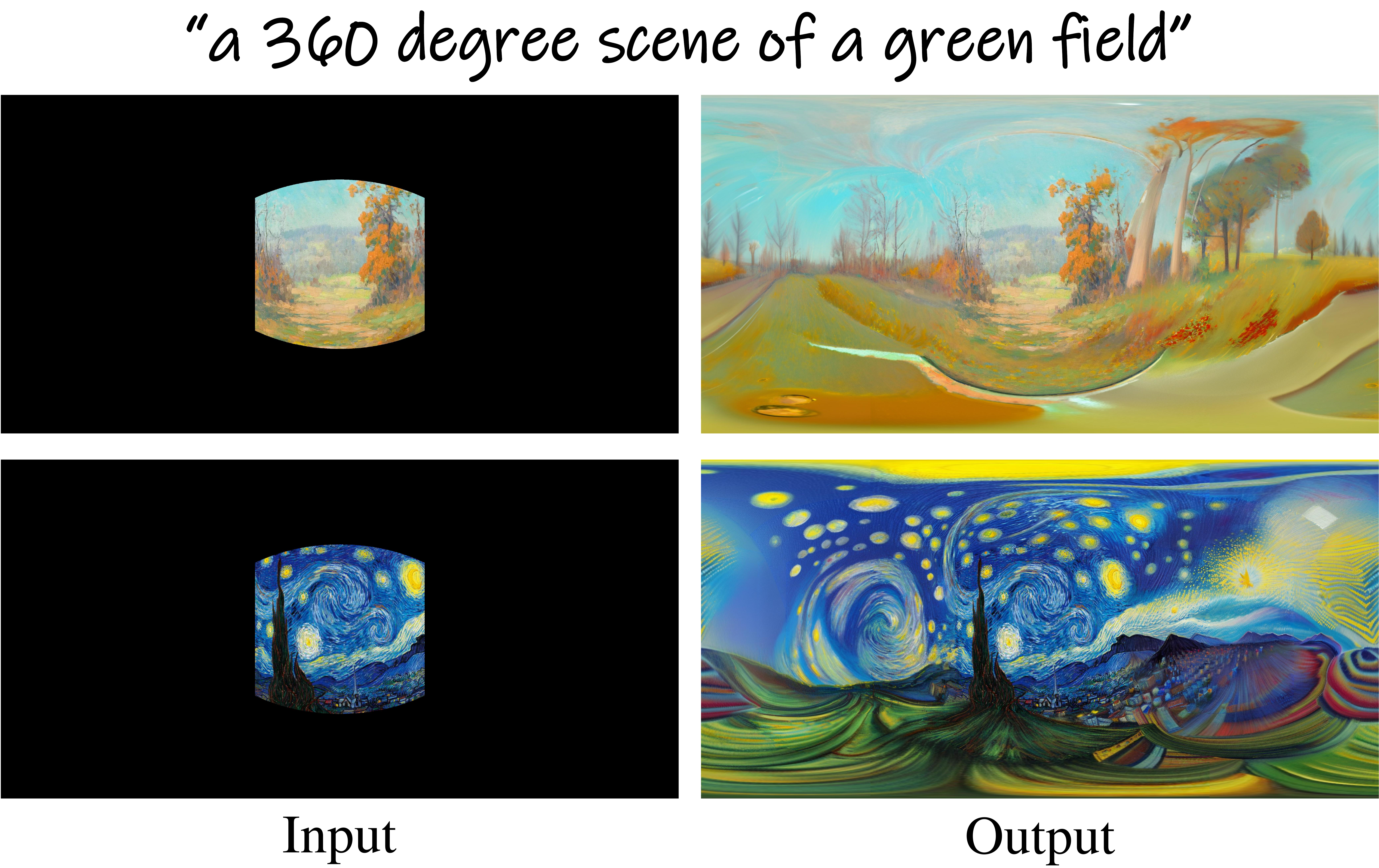}
\caption{Open-image conditioned generation results. }
\label{fig:ood}
\vspace{-0.5em}
\end{figure}

\noindent \textbf{Text-guided Open-image Generation}
In~\autoref{fig:ood}, we present the visual results of open-image generation. The high-quality generated oil-painting images demonstrate that our method could produce vivid pictures with high style consistency for out-of-distribution input, even without style information in the text guidance. This result shows the superiority of the designed VCR, which contributes to keeping the global textual semantic consistency.

\section{Conclusion}
In this paper, we propose a novel 360-degree image out-painting framework with a text-guided diffusion model to generate a consistent and smooth panorama from the NFoV images. We design two novel modules, Visual-textual Consistency Refiner (VCR) and Global-local Mamba Adapter (GMA), to adjust the image-text condition and connect the information flow from the global state to the local adaptation respectively. Extensive experiments show the superiority of our method and achieve state-of-the-art performance.

\bibliographystyle{ACM-Reference-Format}
\bibliography{references}

\end{document}